# MagicVO: End-to-End Monocular Visual Odometry through Deep Bi-directional Recurrent Convolutional Neural Network

Jian Jiao, Jichao Jiao, Yaokai Mo, Weilun Liu, Zhongliang Deng

Beijing University of Posts and Telecommunications
{jiaojian, jiaojichao, moyaokai, liuweilun}@bupt.edu.cn
dengzhl902@gmail.com

## Abstract

*This paper proposes a new framework to solve the problem of monocular visual odometry, called MagicVO . Based on Convolutional Neural Network (CNN) and Bi-directional LSTM (Bi-LSTM), MagicVO outputs a 6-DoF absolute-scale pose at each position of the camera with a sequence of continuous monocular images as input. It not only utilizes the outstanding performance of CNN in image feature processing to extract the rich features of image frames fully but also learns the geometric relationship from image sequences pre and post through Bi-LSTM to get a more accurate prediction. A pipeline of the MagicVO is shown in Fig. 1. The MagicVO system is end-to-end, and the results of experiments on the KITTI dataset and the ETH-asl cla dataset show that MagicVO has a better performance than traditional* visual odometry (VO) systems in the accuracy of pose and the generalization ability.

## 1. Introduction

The problem of estimating ego-motion from a series of consecutive image sequences is a fundamental issue in the robot, which is called visual odometry (VO). VO can be used to estimate the poses of robots and unmanned vehicles by using only cheap cameras. In the past few decades, it has caused widespread concern in the robotics and driverless industries. Among them, feature-based methods and direct methods have achieved great success. However, the feature-based method has the disadvantage of being sensitive to illumination, requiring high image texture and large computational complexity. Moreover, although the direct method is faster, it requires a high sampling rate. In other words, it is required that the object has a small movement to calculate the poses.

In recent years, due to the rapid development of deep learning technologies and the outstanding contributions of CNN in the field of image recognition and segmentation, it has made it possible to use neural networks to deal with VO problems. However, since VO should consider the relevant information of successive image sequences, it needs to process and explore more low-level geometric information among images. Therefore, it is not enough to just use the CNN. Besides, it is not wise to deal with VO problems only through the deep neural network.

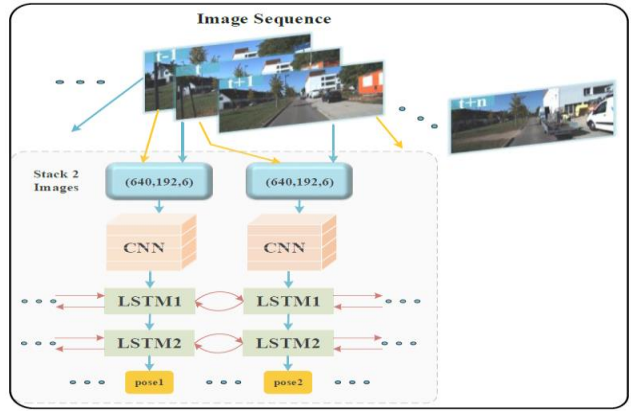

Fig .1. The pipeline of MagicVO system. In the model, two adjacent images are superimposed on the third channel, and the features of image pair are extracted by the CNN, and the geometric relationship of the image sequences pre and post is learned through the Bi-LSTM. Output 6-DoF pose.

In this paper, we propose the MagicVO system, a monocular VO system based on CNN and Bi-LSTM. Our main contributions are as follows:

a) MagicVO system is directly end-to-end and does not require camera internal parameters;

b) The estimated poses are of absolute scale;

c) CNN is used to learn features of image pairs, and the Bi-LSTM is used to learn the relationship between image sequences pre and post, to achieve a high accuracy in VO.

The organization of this article is shown as follows: Section 2 describes the related work, and Section 3 introduces details of our proposed end-to-end MagicVO system. The experimental results are in Section 4. Finally, the conclusion is indicated in Section 5.

## 2. RELATED WORK

Monocular VO has been studied extensively in both robotics and drones. For example, ORB_SLAM2[1] is used by many robots as their own positioning algorithm. In addition, VINS-Mono[2] is a VO algorithm designed for drones. For now, there are two kinds of main methods for this problem: traditional geometric feature-based and learning-based methods.

## 2.1. The method based on traditional geometric features

In general, traditional geometric feature-based methods are implemented using image features and then through rigorous mathematical derivation, where many excellent algorithms are proposed based on traditional geometric features. Therefore, the method occupies the main position of VO.

Furthermore, the traditional geometric feature-based algorithms can be divided into based on sparse features methods and direct methods according to whether images are extracted sparse features and features matching.

1) The sparse feature-based approach has a typical pipeline as shown in Fig.2(a). Based on stereo vision, Nistér et al. (2004) propose one of the earliest VO systems [3]. However, all VO algorithms are reduced sharply in accuracy over time. In order to overcome the problem, visual SLAM is proposed to optimize a more accurate feature map continuously. Early monocular vision SLAM is achieved by means of filters. The extended Kalman filter (EKF) is introduced to achieve simultaneous localization and mapping [4-7]. The main idea of EKF is to use the state vector to store the poses of the camera and the three-dimensional coordinates of the map points. The probability density equation is used to represent the uncertainty in EKF. Moreover, the mean and variance are calculated based on the state of the model to update the state vector. However, the EKF-based SLAM system has the problem of the uncertainty caused by the computational complexity and linearization. In order to compensate for the impact of the linearization, the unscented Kalman filter (UKF) and its related improvements are introduced into the monocular vision SLAM [8-10]. In addition, monocular SLAM is realized by particle filter [11-12]. Although these methods improve the uncertainty, it also increases the computational complexity, which results in decreasing the performance in real-time. To improve the real-time, the monocular vision SLAM based on keyframes has gradually developed. In [13], the authors propose a simple and effective method named PTAM (Parallel Tracking and Mapping) for extracting keyframes. The method improves the real-time to a degree. Since then, the feature-based methods are mostly based on the PTAM model. In [14], the authors present two approaches to improve the agility of a keyframe-based SLAM system: Firstly, they add edge features to the map and exploit their resilience to motion blur to improve the performance of tracking under fast motion. Secondly, the authors implement a very simple inter-frame rotation estimator to aid tracking when the camera moves rapidly. The method improves the robustness of the model when the camera moves quickly. However, the cumulative error is needed to be solved when the camera moves. In [15], the authors present a SLAM method based on keyframe for the relocalization, which can deal with severe viewpoint change for the first time. The method provides an idea for automatic correction of models. In [16], the authors propose a monocular vision SLAM based on keyframes. The whole SLAM process is divided into three threads, which are localization, mapping and loop closing. The method uses ORB features for these three tasks and introduces the concept of Essential Graph to accelerate the correction process of loop closing. This system can run on the CPU in real time, and it can find the original location when it returns to the original scene after dropping frames. However, there are also some problems. For example, it is easy to drop frames when rotating, especially pure rotation. In addition, the point cloud is sparse in the map, so it is not easy to see the specific structure.

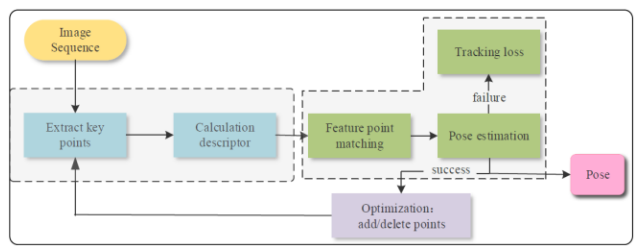

(a). Sparse feature-based method.

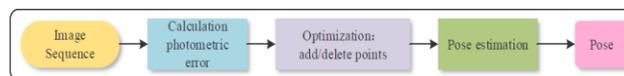

(b). Direct method.

Fig.2. (a). shows a general pipeline based on the sparse feature method, (b). shows the pipeline of the direct method.

In summary, although the feature-based method dominates VO, researchers recognize it with at least the following shortcomings:

a) The extraction of key points and the calculation of descriptors are very time-consuming. If the entire SLAM runs at 30ms/frame, most of the time is spent in calculating feature points.

b) When using feature points, all information except feature points is ignored. An image has hundreds of thousands of pixels, and the feature points are only a few hundred. Most of the image information that may be useful is discarded.

c) Sometimes, cameras move to places where features are not obvious. For example, when we face a white wall or an empty walk. The number of feature points in these scenes will be reduced significantly, we may not find enough matching points to calculate camera motion.

In response to the above shortcomings, the researchers propose a method of calculating the camera motion, which is based on pixels without calculating the key points or calculating the descriptors.

2) The direct method estimates the camera's motion based on the pixel luminance, without having to calculate key points and descriptors at all. Its typical pipeline is shown as Fig.2(b). The direct method is based on the assumption of gray-scale invariant, which has higher accuracy and robustness for environments with fewer features. Monocular visual odometry based on direct method is proposed in recent years. In [17], the authors propose an algorithm of calculating the camera pose by reconstructing a semi-dense inverse depth map. This is the first featureless monocular visual odometry method which runs in real-time on a CPU. However, due to the lack of loop closing detection, the estimated pose uncertainty of this method is large. Moreover, in [18], the authors propose a method for estimating the camera poses by predicting the probabilistic depth for each pixel effectively, which reduces the uncertainty of the pose estimation. However, since this method is based on the assumption of uniformity of light intensity, it is sensitive to the illumination change. In [19], the authors propose a semi-direct monocular visual odometry method of acquiring the camera pose by extracting image blocks, which can enhance the robustness of the algorithm. This approach can reach up to 300 frames/s on a typical laptop. But its accuracy is not high. This is also the result of sacrificing performance at high speeds.

Since the direct method can reduce the time of calculating feature points and descriptors, it has the advantage of high speed. This is one of the reasons why the direct method is becoming more and more popular.

The shortcomings of the direct method are also obvious. The direct method relies on gradient search, which can result the direct method can only succeed when the camera moves very slow. In addition, owing to the direct method calculates the difference between the gradations, the change of overall gradation will destroy the gradation-invariant assumption, causing the algorithm to fail. Therefore, the direct method is sensitive to illumination changes.

## 2.2. Methods based on Deep Learning

Compared with the traditional pose estimation method based on sparse features and dense features, the learning-based method does not require feature extracting and does not require feature matching and complex geometric operations, which makes the deep learning-based methods more intuitive and concise. Konda and Memisevic propose an end-to-end deep neural network architecture for predicting camera speed and direction changes [20]. The main feature of the method is to use a single type of computing module and learning rules to extract visual motion and depth information. According to our experimental results, its proposal proves the feasibility of using deep learning to solve VO problem. In [21], Costante et al. take two consecutive frames as inputs of CNN to learn the optimal image feature representation, which demonstrates the robustness of its algorithm in dealing with image motion blur and illumination changes. Handa et al. extend on the basis of the spatial transform network and choose to regress the classical computer vision method when designing the network, such as end-to-end visual odometry and image depth estimation [22-23]. The authors use the neural network to build a Gvnn (geometric vision with neural network) software library including the global transformation, the pixel transformation, and the M estimator. As an application example, the authors implement a visual odometry based on RGB-D data.

It can be seen from the above that the end-to-end deep neural network architecture can be used to extract inter-frame motion information from the image sequences. Compared with the traditional pose estimation algorithm, the deep learning-based methods replace the cumbersome formula calculation and does not need artificial feature extraction and matching. Therefore, the deep learning-based methods are simple and intuitive. Moreover, pose estimation can be operated online in real-time.

However, those methods ignore considering the relationship between consecutive frame sequences, which result in the accuracy of VO once caught in the bottleneck.

Therefore, considering the fact that Recurrent Neural Network (RNN) can consider the relationship of continuous sequences well, the authors propose a model which connects the CNN with a unidirectional LSTM (DeepVO) [24]. This method can output the relative pose of continuous images. However, the estimation accuracy should be further improved, which is a key factor for the unmanned vehicles.

Based on the method of CNN connecting with unidirectional LSTM, in order to make full use of the information of the consecutive frames, this paper proposes an end-to-end pose estimation that combines CNN and Bi-LSTM. The model does not require camera internal parameters. The experiments have shown that it has a better precision than DeepVO in VO.

## 3. Model for Bidirectional RCNN

In this section, the Bi-RCNN that is an end-to-end model framework for camera pose estimation is described in detail.

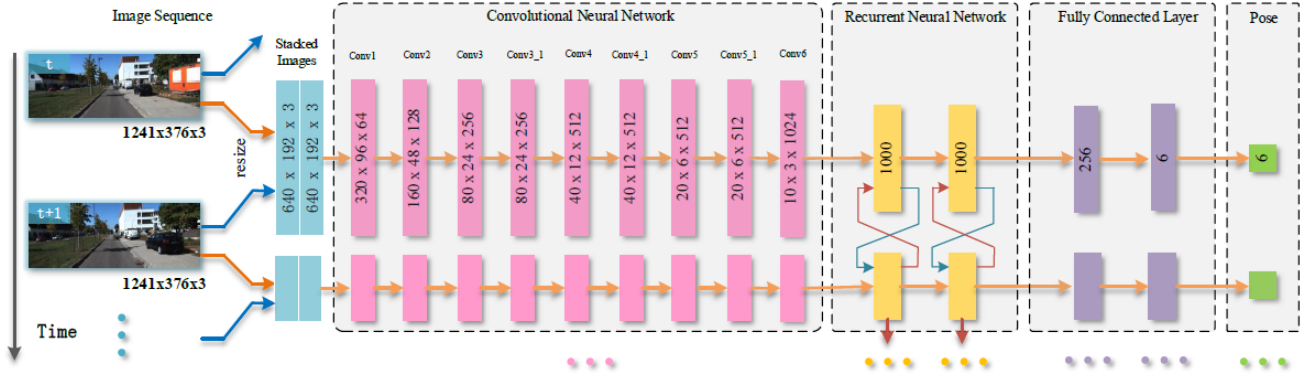

Fig.3. The structure of MagicVO. The parameter from CNN to Bi-LSTM to Fully Connected Layer are shown.

According to Fig.3, we can find that Bi-RCNN includes a CNN structure, an LSTM structure, and two Fully Connected Layers.

### 3.1. Network architecture

The architecture of MagicVO is shown in Fig.3. In the preprocessing section of images, firstly, we resize images to (192,640,3), and we superimpose two adjacent frames on the third channel. Secondly, we subtract the pixel mean and divided by the variance from the superimposed image. The purpose of normalization is to reduce the statistical distribution of the samples and to speed up convergence of model. Then, the CNN takes the preprocessing images as input, and the output of CNN is transmitted to Bi-LSTM as its input. Finally, the 6-DoF pose of each image is output by connecting two Fully Connected Layers. The next part will show more details.

### 3.2. CNN Architecture

There are already many excellent deep networks such as VGGNet, ResNet and SENet for image recognition and segmentation, which mainly extract high-level semantic information of images. However, for VO that relies on more geometric information, there are many problems to solve. Therefore, in order to learn the geometric relationship of images more comprehensively, we refer to the work of [25], which can learn the characteristics of image pairs fully. Table.1 shows the structure of FlowNet. At the same time, in the MagicVO, we also use the network as our part of CNN. In the training of the entire MagicVO, in order to speed up convergence, we use the pre-trained FlowNet network.

### 3.3. Bi-LSTM Architecture

Recently, RNN has been used to solve tasks such as natural language recognition and translation widely, which has achieved good effects.

**Table.1.** The structure of CNN

| Layer   | Kernel Size | Padding | Stride | Number of Channels |
|---------|-------------|---------|--------|--------------------|
| Conv1   | 7×7         | 3       | 2      | 64                 |
| Conv2   | 5×5         | 2       | 2      | 128                |
| Conv3   | 5×5         | 2       | 2      | 256                |
| Conv3_1 | 3×3         | 1       | 1      | 256                |
| Conv4   | 3×3         | 1       | 2      | 512                |
| Conv4_1 | 3×3         | 1       | 1      | 512                |
| Conv5   | 3×3         | 1       | 2      | 512                |
| Conv5_1 | 3×3         | 1       | 1      | 512                |
| Conv6   | 3×3         | 1       | 2      | 1024               |

### 3.4. Bi-LSTM Architecture

Recently, RNN has been used to solve tasks such as natural language recognition and translation widely, which has achieved good effects. The researchers find that RNN has the advantage of limited short-term memory. This is mainly due to the fact that RNN can contact previous information to the current tasks. Given the feature $X_t$ at time $t$, then the RNN is updated at time $t$ by :

$$h_t = f(W_{hh}h_{t-1} + W_{xh}x_t + b_h) \quad (1)$$
$$y_t = W_{hy}h_t + b_y$$

Where $h_t$ is the state variable of the hidden layer at time $t$, and $y_t$ is an output variable. $W_{hh}$ and $W_{xh}$ are the weight matrix of the hidden layers and the input features, respectively, and $b_h$ and $b_y$ are the bias vectors of the hidden layers and the input features, respectively. $f$ is a non-linear activation function, and in general, tanh and sigmoid are used.

However, in the face of long sequence problems, RNN

suffers from severe gradient disappearance in the process of back propagation and gradient descent algorithms, making it difficult for the model to converge to a good effect. The proposal of the LSTM model solves the problem that RNN causes the gradient to disappear on a longer sequence easily. This is mainly due to the addition of three gate systems to the original RNN. However, the LSTM can only have a memory function for the forward sequences, and cannot use backward sequences. Therefore, in order to consider the mutual constraint relationship between image sequences fully, we connect the Bi-LSTM behind CNN. The reason why Bi-LSTM can learn more information about on the sequence pre and post is that it will construct two RNNs forward and backward for each training sequence. Fig.4 shows the structure of Bi-LSTM. Given the feature $x_t$ at time $t$, then the Bi-LSTM is updated at time $t$ by :

$$\begin{aligned} s_t &= f(U x_t + W s_{t-1}) \\ s_t^{'} &= f(U^{'} x_t + W^{'} s_{t+1}^{'}) \\ A_t &= f(W A_{t-1} + U x_t) \\ A_t^{'} &= f(W^{'} A_{t+1}^{'} + U^{'} x_t) \\ y_t &= g(V A_t + V^{'} A_t^{'}) \end{aligned} \quad (2)$$

where $s_t$ and $s_t^{'}$ represent the memory variables of forward and backward at time $t$, respectively. $A_t$ and $A_t^{'}$ are the variables of the hidden layers forward and backward at time $t$, respectively. $y_t$ is an output variable. In addition, $U$, $U^{'}$, $W$, $W^{'}$, $V$ and $V^{'}$ are the weight matrix of the respective variables. $f$ and $g$ are the non-linear activation functions.

Compared with LSTM, Bi-LSTM can further utilize the correlation among the image frames to estimate poses. In our proposed model, we use 1000 LSTM nodes in each direction and a total of 2000 nodes in Bi-LSTM.

### 3.4. Fully-connected layer

Two Fully Connected Layers are connected behind the Bi-LSTM layer to integrate information of the trained image sequences to a final precise pose of 6-DoF, which includes three rotations and three translations. The rotation is represented by Euler angles.

### 3.5. Loss Function

For the proposed MagicVO, it calculates the conditional probability to achieve accurate prediction in the problem of VO. $Y_t=(y_1,y_2,…,y_t)$ is the pose to be predicted, and $X_t=(x_1,x_2,…,x_t)$ is the monocular image sequences, then the conditional probability at time $t$ is shown as follows:

$$p(Y_t | X_t) = p(y_1, y_2, …, y_t | x_1, x_2, …, x_t) \quad (3)$$

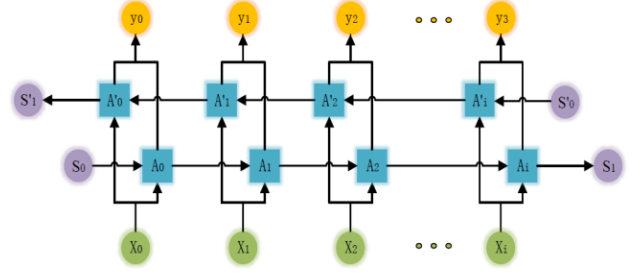

Fig.4. The structure of Bi-LSTM.

We can get the optimal parameter $\theta*$ by maximizing $p(Y_t | X_t)$, as the following equation:

$$\theta^* = \arg\max_\theta P(Y_t | X_t; \theta) \quad (4)$$

In order to get the optimal parameter $\theta^*$, we make the following definition. The translation of output is recorded as $p_k$, and the rotation is expressed with Euler angles vectors denoted as $\varphi_k$. The ground truth of ($p_k$, $\varphi_k$) corresponds to ($p_k$, $\varphi_k$). The loss function is as shown follows:

$$\theta^* = \arg\min_\theta \frac{1}{N} \sum_{i=1}^{N} \sum_{k=1}^{3} \| p_k - p_k \|_2^2 + w \| \varphi_k - \varphi_k \|_2^2 \quad (5)$$

where $\| \cdot \|$ is 2-norm, $k$ is the number of the state quantity in each pose, and because each pose consists of three rotations and three Euler angles, the value of $k$ is from 1 to 3, and $N$ represents the number of image sequences per batch. $w$ is the weight to balance Euler angle and the translation. Finally, the optimal parameter $\theta^*$ can be obtained by performing gradient descent through the neural network iteratively.

## 4. EXPERIMENTAL RESULTS

In this section, we validate our model on the KITTI Visual Odometry / SLAM Evaluation dataset and ETH-asl cla dataset. In order to evaluate our proposed model, the current state-of-the-art feature-based ORB_SLAM2 and VINS-Mono based on optical flow and IMU fusion are compared with MagicOV. It should be noted that in the quantitative analysis, we just use the data in the paper of DeepVO as a comparison, since DeepVO is not an open system, so it is not possible to use the model of DeepVO to have an actual test.

### 4.1. Experimental Setup

This network model is trained with the PyTorch-0.4.1 framework in Intel(R) Xeon(R) CPU E5-2630 v3 @

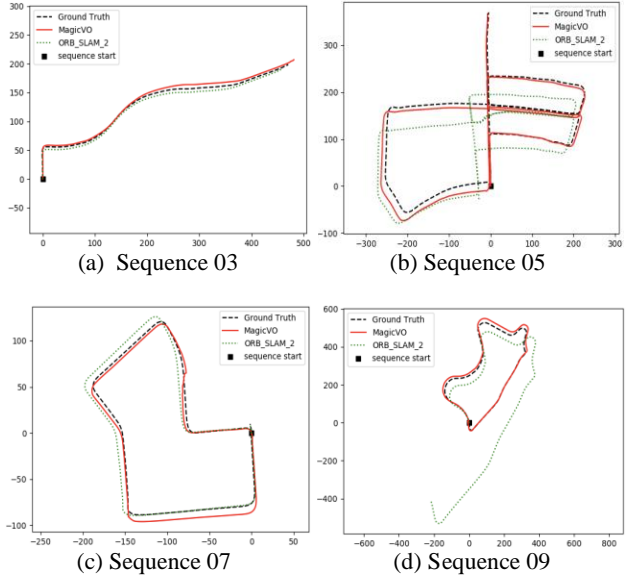

(a) Sequence 03  (b) Sequence 05

(c) Sequence 07  (d) Sequence 09

Fig.5. MagicVO trained on the sequence 00, 02, 04, 06, 08, 10, trajectory verified on the 03, 05, 07, 09 sequence.

**Table 2.** Comparison results on the verification set.

| Seq. | MagicVO | | ORB_SLAM2 | | DeepVO | |
|---|---|---|---|---|---|---|
| | t(%) | r(°) | t(%) | r(°) | t(%) | r(°) |
| 03 | **4.95** | 2.44 | 10.75 | **2.34** | 8.49 | 6.89 |
| 05 | **1.63** | **2.25** | 24.66 | 3.45 | 2.62 | 3.61 |
| 07 | **2.61** | **1.08** | 3.81 | 1.67 | 3.91 | 4.60 |
| 09 | **5.43** | 3.31 | 78.75 | **1.78** | \ | \ |
| mean | **3.66** | **2.27** | 29.49 | 2.31 | 5.01 | 5.03 |

t: The RMSE of translation on KITTI dataset.
r: The RMSE of rotation on KITTI dataset.
\: The part of data is not given in the paper (DeepVO).

2.40GHz, GPU is Nvidia titan X pascal. RAM is 64G and Graphics Card Ram Size is 12G.

### 4.2. Dataset

The KITTI Visual Odometry / SLAM Evaluation benchmark consists of 22 stereo sequences, where each image is saved as png format. It provides 11 sequences (NO.00-NO.10) with ground truth trajectories for training and verification. The other 11 sequences (NO.11-NO.21) without ground truth are utilized for testing our proposed MagicVO. Moreover, the data set is sampled at 10fps at an average speed of 90km/h, so this is a big challenge to use the dataset for training and testing our model.

In addition, in order to further verify the generalization of MagicVO, we also compare with the VINS-Mono on the ETH-asl cla dataset. The dataset is collected by the drone completely. There are four sequences, cla_floor_f, cla_floor_f_v2, cla_floor_g and cla_floor_j. The jitter of the dataset is very severe, therefore, there is a big challenge for the model. Examples from KITTI and ETH data sets are shown in Fig. 8.

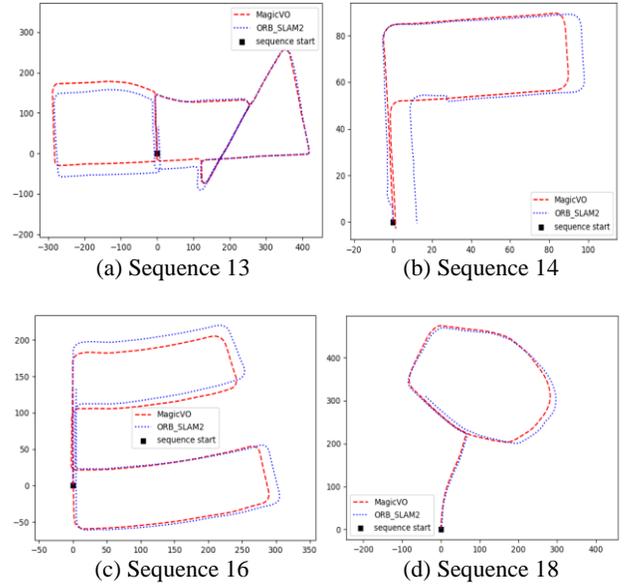

(a) Sequence 13  (b) Sequence 14

(c) Sequence 16  (d) Sequence 18

Fig.6. The comparison between MagicVO and ORB_SLAM2 on the KITTI test set. The model is trained on Sequence 00, 02, 04, 06, 08, 10.

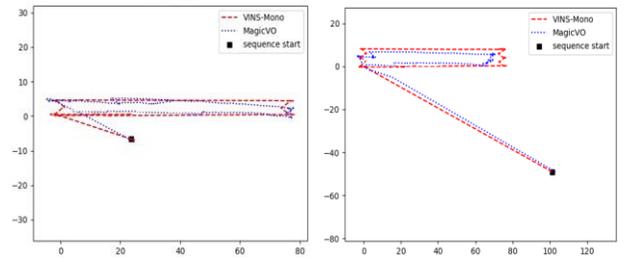

Fig.7. Trajectories of MagicVO and VINS-Mono on cla_floor_f and cla_floor_g of ETH-asl cla dataset. The models are trained on cla_floor_f_v2_for_localization and cla_floor_j of ETH-asl cla dataset.

It is noted that we have also tried to use the ORB_SLAM2 on ETH-asl cla as a comparison, but since the jitter is so severe that the ORB_SLAM2 cannot work properly.

### 4.3. Training and Testing

This part introduces some details in training, verification and testing of MagicVO.

**4.3.1 Training skills**. In order to speed up training time and reduce the amount of data required, we use the pre-trained the FlowNet model.

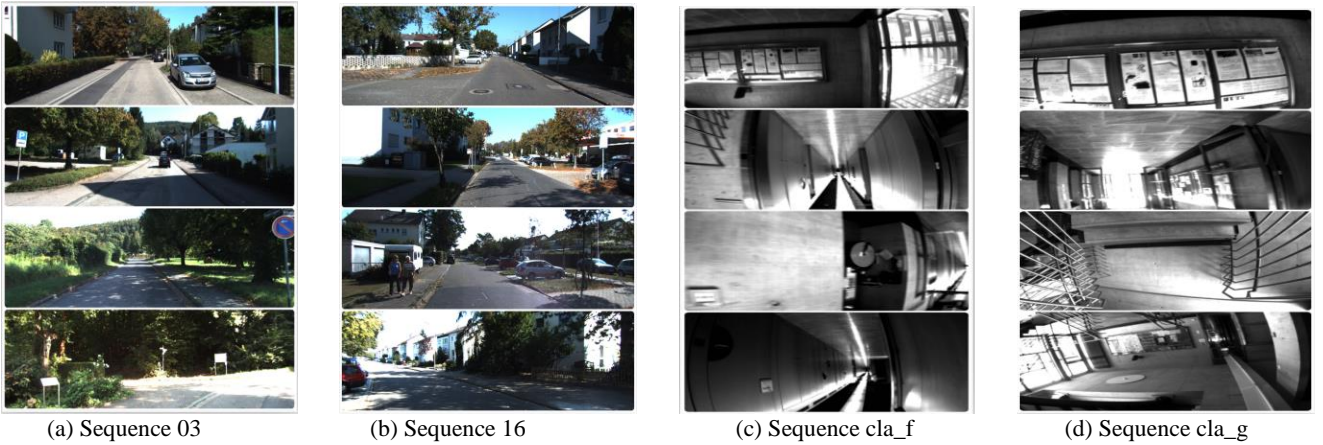

(a) Sequence 03   (b) Sequence 16   (c) Sequence cla_f   (d) Sequence cla_g

Fig.8. Sequence 03, 16 of KITTI dataset and Sequence cla_f , cla_g of ETH-asl cla dataset with 4 sample images for each sequence.

In addition, in order to be able to meet Graphics Card Ram Size, we reduce the image to 1/4 of the original image, which can accelerate the convergence of the model and achieve the expected effect finally.

In the selection of model parameters, we choose the Adagrad optimizer and the learning rate is set to 0.001. In order to prevent the model from overfitting, we use the way of Dropout among the Bi-LSTM layers. In order to prevent the model from gradient explosions, we also use gradient clipping in Bi-LSTM.

**4.3.2 Verification and testing.** In order to ensure that our model has not been overfitting, we use Sequence 03, 05, 07,and 09 as the verification set. Furthermore, we test our proposed model on Sequence 13, 14, 16, 18 to prove that our model has a good generalization. Moreover, we also train and test our model on the ETH-asl cla dataset, which also achieve good results.

Since the KITTI dataset only has a total of 11 sequences to provide ground truth. Therefore, we use the Sequence 00, 02, 04, 06, 08 and 10 for training, the Sequence 03, 05, 07 and 09 for verification, and the remaining Sequence 11-21 for testing. In addition, to further verify the generalization of MagicVO, we train our model on the cla_floor_f_v2 and cla_floor_j sequences of the ETH-asl cla dataset and test the model on cla_floor_f and cla_floor_g.
The proposed method is mainly measured in three ways.

Firstly, the RMSE of the model is calculated in terms of rotation and translation. Our model is trained on Sequence 00, 02, 04, 06, 08, 10, verified on 03, 05, 07, 09. It is compared with ORB_SLAM2 in the same sequences. Fig.5 shows the trajectory on the verification set. Moreover,

Table2 shows the error comparison between our model and ORB_SLAM2.

Secondly, we test the model on Sequence 13, 14, 16, 18, which are compared with ORB_SLAM2. The comparison results are shown in Fig.6.

Finally, to further verify the generalization of the model, we test the model on ETH-asl cla dataset. The trajectories are shown in Fig.7.

**4.4. Result analysis**

In this part, we mainly analyze the reasons why several models have different performance on KITTI and ETH-asl cla datasets.

**4.4.1 For the KITTI data set.** In general, we find that MagicVO performs well from Table.1, Fig.5 and Fig.6, especially on translation, but ORB_SLAM2 is not effective in calculating the translation. We think there are two main reasons for the result:

a) Since the image of the KITTI dataset is collected by the car at a higher speed, so the movement of adjacent frames is relatively large, which leads that feature points used to match between adjacent frames are reduced. Therefore, the feature-based ORB_SLAM2 does not work well when calculating the rotation. However, with the Bi-LSTM, MagicVO can make more use of the information of the pre and post frames extracted through CNN. This is one of the reasons why MagicVO performs well.

b) From Fig.8 (a) and (b), we can see that there are dynamic objects in some scenes, such as the second picture of sequence 03, the third picture of sequence 16. If the dynamic object has a large proportion in a picture, it will have a serious impact on feature-based algorithms. Because if the feature points in the previous picture are concentrated on the dynamic object, and the dynamic object has a large movement in the next adjacent frames, which will have a serious effect on the feature matching and cause a large error. The feature extraction of MagicVO is based on CNN, compared with artificial features, such as ORB, it has the advantage of extracting deeper feature information to making it insensitive to dynamic objects. This is another important reason why MagicVO performs well on KITTI.

**4.4.2 For the ETH-asl cla dataset.** From Fig.7, we find that the performance of Magic on the ETH-asl cla dataset is worse than that on KITTI, but VINS-Mono has better performance. After comparing the two datasets carefully, and analyzing the characteristics of the two models, we summarize that there are two reasons for the result:

a) The number of images used for training is limited. Since we only use two image sequences for model training, and the number of images in each image sequence is only about 3500, so the model cannot learn the rules well. I think that is one of the reasons why MagicVO performs relatively badly. We think if we can get more training data, Magic can perform better. However, the optical flow-based method does not require a large amount of data.

b) Since the ETH-asl cla dataset is collected by drones, the jitter is very intense, and there are many blurry frames, which increases the noise during estimating poses, shown as Fig.8 (c) and (d). Even we have tried to run ORB_SLAM2 on the dataset, but it does not work. This is another major reason why MagicVO does not achieve better results. In the case, VINS-Mono can make more use of the information that is provided by IMU to correct the error caused by the picture information.

## 5.Conclusion

We propose a new end-to-end deep learning framework MagicVO that calculates camera poses on a sequence of continuous monocular images. MagicVO is composed of CNN and Bi-LSTM. After comparing to the other two states of art algorithms on KITTI and ETH data sets, the results showed that our proposed algorithm can achieve an outperformance in pose estimation accuracy and illumination invariant. In addition, our method allows the camera to be performed at the high speed that is no less than 90 KM/H.

In the future work, we will focus on unsupervised learning algorithms for reducing the number of training data. Although in some scenarios our method is comparable to the traditional SLAM method, our method needs a lot of data to be trained with ground truth, which is the shortcoming of the deep learning based methods.

**References**

[1] R. Mur-Artal, J. M. M. Montiel, and J. D. Tardos, "ORB-SLAM: A Versatile and Accurate Monocular SLAM System," *IEEE Trans. Robot.*, vol. 31, no. 5, pp. 1147–1163, 2015.

[2] T. Qin, P. Li, and S. Shen, "VINS-Mono : A Robust and Versatile Monocular Visual-Inertial State Estimator," *IEEE Trans. Robot.*, vol. 34, no. 4, pp. 1004–1020, 2018.

[3] D. Nistér, O. Naroditsky, and J. Bergen, "Visual odometry," *Proc. 2004 IEEE Comput. Soc. Conf. Comput. Vis. Pattern Recognit. 2004 CVPR 2004*, vol. 1, no. C, pp. 652–659, 2004.

[4] a J. Davison, "SLAM with a Single Camera," *SLAMCML Work. ICRA 2002*, 2002.

[5] A. J. Davison, "Real-time Simultaneous Localisation and Mapping with a Single Camera," *Iccv*, vol. 2, pp. 1403–1410, 2003.

[6] A. J. Davison, I. D. Reid, N. D. Molton, and O. Stasse, "MonoSLAM: Real-time single camera SLAM," *IEEE Trans. Pattern Anal. Mach. Intell.*, vol. 29, no. 6, pp. 1052–1067, 2007.

[7] J. Civera, A. J. Davison, and J. M. M. Montiel, "Inverse Depth Parameterization for Monocular {SLAM}," *IEEE Trans. Robot.*, vol. 24, no. 5, pp. 932–945, 2008.

[8] R. Martinez-cantin, "Unscented SLAM for Large-Scale Outdoor Environments.pdf."

[9] D. Chekhlov, M. Pupilli, W. Mayol-cuevas, and A. Calway, "Real-Time and Robust Monocular SLAM Using Predictive Multi-resolution Descriptors," *Adv. Vis. Comput. Second Int. Symp. ISVC 2006 Lake Tahoe, NV, USA, Novemb. 6-8, 2006. Proceedings, Part II*, vol. 4292/2006, pp. 276–285, 2006.

[10] S. Holmes, G. Klein, and D. W. Murray, "A Square Root Unscented Kalman Filter for visual monoSLAM," *Proc. - IEEE Int. Conf. Robot. Autom.*, pp. 3710–3716, 2008.

[11] R. Sim, P. Elinas, M. Griffin, and J. J. Little, "Vision-based SLAM using the Rao-Blackwellised Particle Filter," *Proc. IEEE*, vol. 14, no. July, pp. 9–16, 2005.

[12] M. Li, B. Hong, Z. Cai, and R. Luo, "Novel Rao-Blackwellized particle filter for mobile robot SLAM using monocular vision," *Int. J. Intell. …*, vol. 1, no. 1, pp. 1021–1027, 2006.

[13] G. Klein and D. Murray, "Parallel tracking and mapping for small AR workspaces," *2007 6th IEEE ACM Int. Symp. Mix. Augment. Reality, ISMAR*, 2007.

[14] G. Klein and D. Murray, "Improving the agility of keyframe-based SLAM," *Lect. Notes Comput. Sci. (including Subser. Lect. Notes Artif. Intell. Lect. Notes Bioinformatics)*, vol. 5303 LNCS, no. PART 2, pp. 802–815, 2008.

[15] R. Mur-Artal and J. D. Tardós, "Fast relocalisation and loop closing in keyframe-based SLAM," *Proc. - IEEE Int. Conf. Robot. Autom.*, pp. 846–853, 2014.

[16] R. Mur-Artal, J. M. M. Montiel, and J. D. Tardos, "ORB-SLAM: A Versatile and Accurate Monocular SLAM System," *IEEE Trans. Robot.*, vol. 31, no. 5, pp. 1147–1163, 2015.

[17] J. Engel, J. Sturm, and D. Cremers, "Semi-dense visual odometry for a monocular camera," *Proc. IEEE Int. Conf. Comput. Vis.*, pp. 1449–1456, 2013.

[18] R. A. Newcombe, S. J. Lovegrove, and A. J. Davison, "DTAM :Dense Tracking and Mapping in Real-Time " pp. 2320–2327, 2011.

[19] C. Forster, M. Pizzoli, and D. Scaramuzza, "SVO: Fast semi-direct monocular visual odometry," *Proc. - IEEE Int. Conf. Robot. Autom.*, 2014.

[20] K. Konda and R. Memisevic, "Learning Visual Odometry with a Convolutional Network," *Int. Conf. Comput. Vis. Theory Appl.*, pp. 486–490, 2015.

[21] G. Costante, M. Mancini, P. Valigi, and T. A. Ciarfuglia, "Exploring Representation Learning With CNNs for Frame-to-Frame Ego-Motion Estimation," *IEEE Robot. Autom. Lett.*, vol. 1, no. 1, pp. 18–25, 2016.

[22] A. Handa, M. Bloesch, V. Pătrăucean, S. Stent, J. McCormac, and A. Davison, "Gvnn: Neural network library for geometric computer vision," *Lect. Notes Comput. Sci. (including Subser. Lect. Notes Artif. Intell. Lect. Notes Bioinformatics)*, vol. 9915 LNCS, pp. 67–82, 2016.

[23] S. K. Sønderby, C. K. Sønderby, L. Maaløe, and O. Winther, "Recurrent Spatial Transformer Networks," pp. 1–9, 2015.

[24] S. Wang, R. Clark, H. Wen, and N. Trigoni, "DeepVO: Towards End to End Visual Odometry with Deep Recurrent Convolutional Neural Networks," *IEEE Int. Conf. Robot. Autom.*, pp. 2043–2050, 2017.

[25] A. Dosovitskiy *et al.*, "{FlowNet}: {L}earning Optical Flow with Convolutional Networks," *{IEEE} Int. Conf. Comput. Vis.*, pp. 2758–2766, 2015.